\documentclass{article}
\usepackage[preprint]{neurips_2025}
\usepackage[linesnumbered,ruled,vlined]{algorithm2e}
\usepackage{enumitem}
\usepackage{arydshln} 
\usepackage{amssymb}
\usepackage[utf8]{inputenc} 
\usepackage[T1]{fontenc}    
\usepackage{hyperref}       
\usepackage{url}            
\usepackage{booktabs}       
\usepackage{amsfonts}       
\usepackage{nicefrac}       
\usepackage{microtype}      
\usepackage[table]{xcolor}          
\usepackage{graphicx}
\usepackage{booktabs}
\usepackage{multirow}
\usepackage{bbding}
\usepackage{amsmath}
\usepackage{orcidlink}

\usepackage{algorithmic}
\usepackage[capitalize,noabbrev,nameinlink]{cleveref}
\usepackage{subfigure}
\usepackage{makecell}
\usepackage{subcaption}  
\usepackage{graphicx}
\usepackage{lipsum}    
\usepackage{wrapfig}
\DeclareMathOperator*{\argmax}{arg\,max}
\usepackage[misc]{ifsym}
\title{Training-Free Watermarking for Autoregressive\\ Image Generation}

\makeatletter
\def\thanks#1{%
  \protected@xdef\@thanks{\@thanks \protect\footnotetext[0]{#1}}
}
\makeatother

\author{%
  Yu Tong$^{1, 2}$\quad 
  Zihao Pan$^{3}$\quad
  Shuai Yang$^{4}$\quad
  Kaiyang Zhou$^{1, \textrm{\Letter}}$\thanks{$^\textrm{\Letter}~$Corresponding author}\\
  $^{1}$Hong Kong Baptist University\quad
  $^{2}$Wuhan University\\
$^{3}$Sun Yat-sen University\quad
$^{4}$Peking University\\
\href{https://github.com/maifoundations/IndexMark}{https://github.com/maifoundations/IndexMark}
 }

\begin{document}

\maketitle
\begin{abstract}
Invisible image watermarking can protect image ownership and prevent malicious misuse of visual generative models. However, existing generative watermarking methods are mainly designed for diffusion models while watermarking for autoregressive image generation models remains largely underexplored. We propose IndexMark, a training-free watermarking framework for autoregressive image generation models. IndexMark is inspired by the redundancy property of the codebook: replacing autoregressively generated indices with similar indices produces negligible visual differences. The core component in IndexMark is a simple yet effective match-then-replace method, which carefully selects watermark tokens from the codebook based on token similarity, and promotes the use of watermark tokens through token replacement, thereby embedding the watermark without affecting the image quality. Watermark verification is achieved by calculating the proportion of watermark tokens in generated images, with precision further improved by an Index Encoder. Furthermore, we introduce an auxiliary validation scheme to enhance robustness against cropping attacks. Experiments demonstrate that IndexMark achieves state-of-the-art performance in terms of image quality and verification accuracy, and exhibits robustness against various perturbations, including cropping, noises, Gaussian blur, random erasing, color jittering, and JPEG compression.
\end{abstract}
\section{Introduction}
With the remarkable success of Large Language Models (LLMs)~\citep{attention,brown2020language,zhang2022opt} in natural language processing, recent advancements have seen autoregressive image generation models, such as LlamaGen~\citep{llamagen} and VAR~\citep{var}, demonstrating substantial potential in the domain of visual generation. These models leverage a Vector Quantization (VQ) tokenizer~\citep{van2017neural} to transform images into discrete tokens. Subsequently, they autoregressively predict the ``next token'' within a codebook to generate images. Notably, these models exhibit significant advantages in terms of both image quality and generation speed. The proliferation of open-source high-quality autoregressive image generation models empowers the general public to create diverse customized content. However, it also brings potential risks of model misuse~\citep{brundage2018malicious,zohny2023ethics,vincent2020online}, such as fake news fabrication, ambiguous copyright attribution, and improper use of public figures' portraits. Amidst growing calls for government regulation and industry compliance~\citep{whitehouse,microsoft}, model developers need to enhance image traceability to ensure accountability in legal liability determination, copyright protection, and content moderation.

Invisible watermarking~\citep{robin,dwtdct,lawa} provides a technical pathway for image traceability. This technology embeds imperceptible watermarks into images to help model developers achieve user-level attribution tracking of AI-generated content. Existing watermarking methods can be broadly categorized into two types: post-processing watermarks embedded after generation~\citep{dct,dwt,dwtdct}, and generative watermarks integrated during the generation process~\citep{yu2020responsible,fernandez2023stable}. Since the former introduces additional inference and storage overhead, generative watermarks are generally more practical and hence more popular. However, current generative watermarking techniques primarily focus on diffusion models and lack exploration for the emerging autoregressive image generation models. Due to substantial architectural differences between the two paradigms---diffusion models employ progressive denoising~\citep{ddpm} whereas autoregressive models rely on sequential generation~\citep{llamagen}---current diffusion-based watermarking methods cannot be directly applied to autoregressive models.
This motivates us to investigate efficient and effective autoregressive watermarking strategies.

\begin{wrapfigure}[18]{r}{0.4\textwidth}
\centering
\vspace{-1em}
\includegraphics[width=0.4\textwidth]{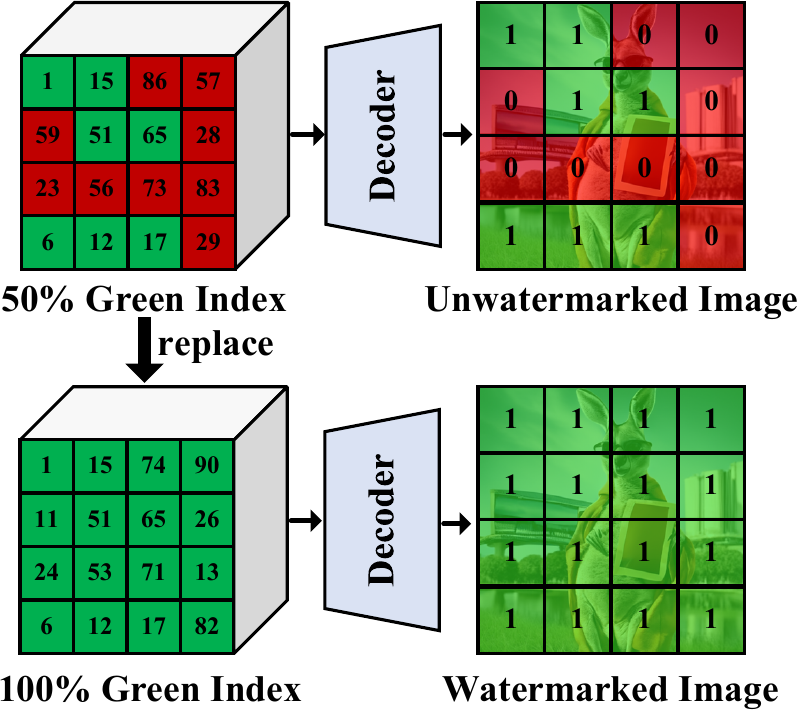}
\caption{\small Watermark embedding by index replacement to attain a higher proportion of watermark tokens (green index).}
\label{fig:introduce}
\end{wrapfigure}

We believe \textit{leveraging the characteristics of the autoregressive image generation models is the key to the design of an effective watermarking strategy}. 
Recent research in autoregressive image generation models has identified a notable \textit{redundancy} issue in their codebooks~\citep{hu2025codebook,guo2025codebook}: a large number of vectors are associated with different indices but highly similar to each other. This feature naturally leads to an elegant solution of watermarking that has minimal impact on the content of the image. Instead of using an image as a watermark, we embed the watermark by altering the statistical distribution of generated indices~\citep{kirchenbauer2023watermark}. Specifically, we divide the codebook into ``red'' and ``green'' groups by pairing similar indices. After generating the indices, we replace as many red indices as possible with their paired green indices (called watermark tokens), thus changing the distribution ratio of red-green indices in the final sequence to embed watermark information (see Figure~\ref{fig:introduce}). This type of watermark has three advantages. \textbf{1)} It is inherently robust and can only be removed by extensively modifying the color blocks of the image. \textbf{2)} The redundancy of the codebook allows this \textit{match-then-replace} strategy to imperceptibly embed watermarks. 
\textbf{3)} Moreover, different red-green division schemes can correspond to different identity identifiers (IDs), assisting model developers
in image tracing. 

Building on the above insights, this paper proposes IndexMark, the first $\textit{training-free}$ watermarking framework for autoregressive image generation models. We first formulate the pairing of similar indices as a $\textit{maximum weight perfect matching}$ problem~\citep{maxperfect}, and solve it with top-K pruning and the Blossom algorithm~\citep{blossom}. Then, we randomly assign each pair of indices to a red list or a green list. After autoregressive index generation, red indices are selectively replaced with their paired green indices according to index confidence, thereby embedding an invisible image watermark. This \textit{match-then-replace} strategy robustly embeds watermarks with image quality and content well preserved. 
During the watermark verification stage, indices of the generated image are reconstructed via VQ-VAE to compute the ``green-index rate'' for verification. To compensate the index reconstruction errors of VQ-VAE, we introduce an Index Encoder for accurate index reconstruction. Although the red-green watermark is intrinsically robust against various image perturbations, the verifier is still vulnerable to cropping attacks due to VQ-VAE's block-level processing characteristics. Therefore, we propose a corresponding cropping-robust validation scheme specific to the modern autoregressive image generation models.

Our key contributions are summarized as follows:
1) We propose a training-free watermarking framework that can be directly applied to autoregressive image generation models without requiring any additional fine-tuning or training.
2) We introduce a match-then-replace approach, which enables training-free watermark embedding with minimal impact on the visual quality of the images.
3) We design a precise image indexing validation framework that can verify the presence of watermarks with higher statistical confidence.
4) Thanks to the Index Encoder and the designed cropping-robust watermark verification method, our approach demonstrates strong robustness towards a wide range of image perturbations.

\section{Related Work}
\subsection{Image Watermarking}
Watermarks in generative models can be embedded either after images are generated (\textit{i.e.}, post-processing) or during the generation process (\textit{i.e.}, in-processing). Post-processing methods are mainly divided into transformation-based and deep encoder-decoder methods. Representative post-processing methods include Discrete Wavelet Transform (DWT)~\citep{dwt}, Discrete Cosine Transform (DCT)~\citep{dct}, and DWT-DCT methods~\citep{dwtdct}, which embed watermarks into the spatial or frequency domain with minimal impact on image quality. A drawback of these methods is that simple attacks in the pixel space can significantly affect the accuracy of watermark extraction. Deep encoder-decoder methods often generate watermarked images in an end-to-end manner, \textit{e.g.}, using adversarial training to add imperceptible noise into image pixels~\citep{hidden, RivaGAN}. However, these methods often struggle to generalize to images outside the training data distribution.

Research on in-processing watermarking primarily focuses on diffusion-based models. The Tree-Ring watermark~\citep{tree} embeds watermark into the noisy image before denoising. ROBIN~\citep{robin} injects watermark into an intermediate diffusion state while maintaining consistency between the watermarked image and the generated image and robustness of the watermark. Though the performance is strong, these methods cannot be directly transferred to autoregressive architectures.
Concurrent to our work, Safe-VAR~\citep{safevar} explores watermark embedding for the VAR model~\citep{var} through multi-scale interaction and fusion techniques. However, Safe-VAR lacks scalability as it only works for VAR-like models and suffers from high training costs as it requires over 200K images for fine-tuning the decoder and training an additional multi-scale module. In contrast, our approach does not require any training and can be applied more broadly to codebook-based autoregressive models.

\subsection{Autoregressive Image Generation}
In autoregressive image generation, image data is typically transformed into one-dimensional sequences of pixels or tokens, and the model predicts the next image token based on the existing context. 
Early autoregressive image generation models~\citep{pixelcnn,pixelrnn} perform image generation by predicting continuous pixels, which have high computational complexity.
The seminal work, VQ-VAE~\citep{van2017neural}, builds a codebook containing feature representations and casts image generation into a discrete label prediction problem. VQ-GAN~\citep{esser2021taming} extends VQ-VAE by using adversarial training to improve the image quality.
Recently, LlamaGen~\citep{sun2024autoregressive} and Open-MAGVIT2~\citep{luo2024open} apply the concept of next-token prediction, which has been widely used in large language models, to autoregressive image generation, achieving performance that even surpass diffusion-based methods.\footnote{The background on autoregressive image generation can be found in the Appendix~\ref{vqvaear}.}
However, the problem of embedding watermarks into autoregressive image generation models remains largely understudied, exposing huge risks of model misuse for these models. Our IndexMark fills this gap by offering a simple, training-free solution.

\section{Methodology}\label{sec:Methodology}
\paragraph{Task Definition} 
Autoregressive model watermarking aims to embed an invisible, verifiable watermark $w$ into an autoregressively generated image $I$ during creation using a watermarking algorithm $\mathcal{E}$, resulting in $I_w$. After potential real-world image transformations $\mathcal{O}$ such as JPEG compression, the model owner can extract the watermark from the altered image $\mathcal{O}(I_w)$ via a validation algorithm $\mathcal{D}$, facilitating image traceability. 

\begin{figure*}[t]
    \centering
    \includegraphics[width=\linewidth]{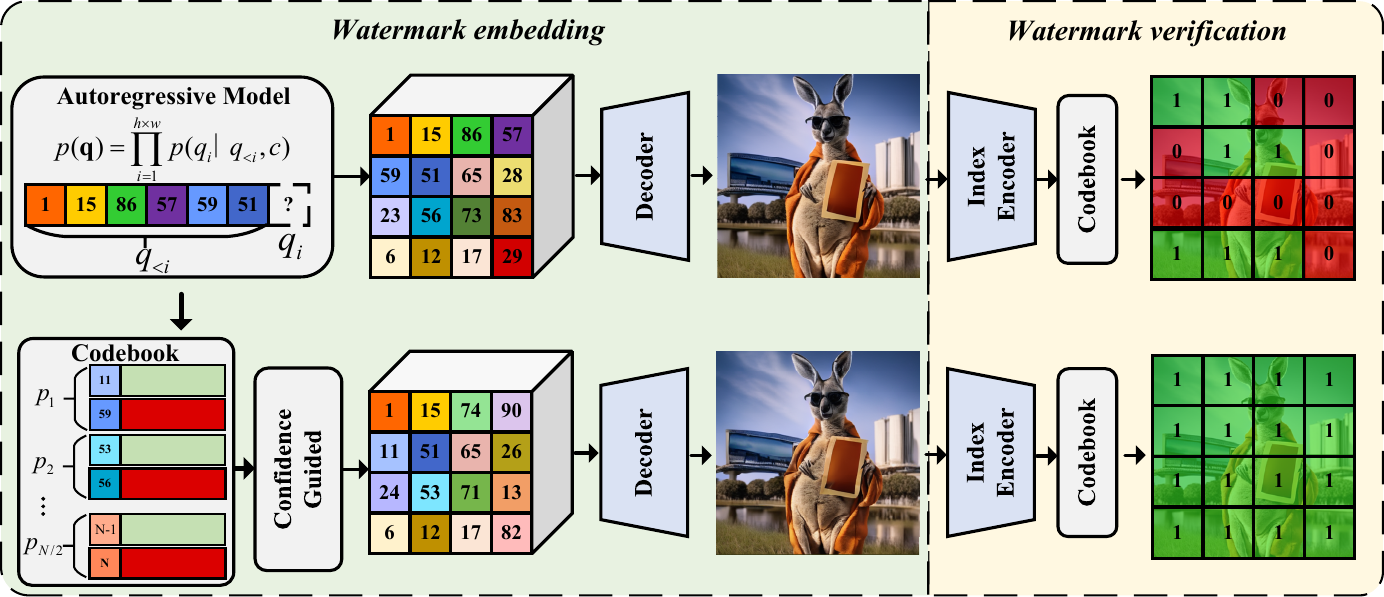}
    \caption{Watermark embedding and verification of IndexMark. During autoregressive index generation, IndexMark selectively replaces red indices with green indices from the same index pair based on confidence to embed the watermark. The watermarked image is fed into the Index Encoder to calculate the green index rate for watermark verification.}
    \label{fig:model}
    \vspace{-15pt}
\end{figure*}

\paragraph{Our Framework} As illustrated in Figure~\ref{fig:model}, our IndexMark framework is composed of two parts: watermark embedding (Sec.~\ref{Watermark Embedding}) and watermark verification (Sec.~\ref{sec:Watermark Verification}). In the watermark embedding part, we first divide the codebook of an autoregressive model into pairs of indices such that each pair contains similar vectors. Then, for each index pair we randomly assign one index into a red list and the other into a green list. Finally, we perform selective red-green index replacement based on index confidence during the image decoding process to embed invisible watermarks into images (\textit{i.e.}, replacing as many red indices as possible with green indices). In the watermark verification part, we propose a method based on statistical probability, where an Index Encoder is introduced to achieve precise reconstruction of indices. We also propose a cropping-invariant watermark verification scheme for cropped images.

\subsection{Watermark Embedding}\label{Watermark Embedding}

\paragraph{Construction of Index Pairs} \label{Construction}
We aim to divide the indices in the codebook, $\mathcal{I}=\{1,2,...,N\}$, into index pairs $\mathcal{P}=\{p_1,p_2,...,p_{N/2}\}$, where each pair $p_k = \{i_k, j_k\}$ contains two distinct indices from $\mathcal{I}$ such that $\bigcup_{k=1}^{N/2} p_k = \mathcal{I}$ and $p_k \cap p_{k'} = \varnothing$ for $k \neq k'$. 
The objective of the red-green index allocation is to compute an optimal assignment that maximizes the sum of the intra-pair similarity $S_{sum}$ for all red-green index pairs:
\begin{equation}
    S_{sum} = \max_{\mathcal{P} \in \mathbb{P}} \sum_{k=1}^{N/2} \text{sim}(i_k, j_k),
\end{equation}
where $\mathbb{P}$ is the set of all possible such partitions $\mathcal{P}$ and $\text{sim}(i, j)$ is the cosine similarity between the vectors of index $i$ and  index $j$ in the codebook. We cast this problem into a \textit{maximum weight perfect matching} problem. Specifically, we construct a complete graph $G = (V,E)$, where each vertex in the vertex set $V$ represents an index from the codebook and the edge set $E$ connects all pairs of vertices, with the edge weights $w$ set as the cosine similarity between the two connected vertices. After constructing this complete graph, our objective is to find a maximum weight perfect matching $M^*$, which is a subset of $E$ containing $N/2$ edges such that every vertex in $V$ is linked to only one edge in $M^*$, and the sum of the weights of these $N/2$ edges is maximized:
\begin{equation}
    M^* = \argmax_{M \in \mathbb{M}} \sum_{(i,j) \in M} w(i, j),
\end{equation}
where $\mathbb{M}$ represents the set of all possible perfect matchings on the graph $G$, and $w(i,j)$ represents the weight of the edge connecting vertex $i$ and vertex $j$.

We solve this problem using the Blossom algorithm~\citep{blossom}. Considering the large number of indices in the codebook, directly applying the Blossom algorithm would result in extremely high computational complexity. For this reason, we perform top-K pruning on the complete graph, retaining only the $K$ edges with the highest weights for each vertex. We then apply the Blossom algorithm~\citep{blossom} to the pruned sparse graph to obtain the maximum weight perfect matching $M^*$. 
Please refer to Appendix~\ref{blossomapp} for the details on the Blossom algorithm.

\paragraph{Red-Green Index Assignment}
\label{Assignment}
After obtaining the maximum weight perfect matching $M^*$, we need to assign indices to red and green lists for each index pair in $M^*$. For simplicity, we randomly assign the two indices in each index pair to the red and green lists. In practical applications, users can customize the assignment of red and green indices. The total number of possible assignments is as high as $2^{N/2}$, providing model developers and users with extremely abundant identity identifiers (IDs) for image tracing.

\paragraph{Confidence-Guided Index Replacement} 
Autoregressive image generation models produce token index sequences in an autoregressive manner. Our objective is to replace as many red indices as possible with green indices, while avoiding bad replacements that harms image quality.~
To achieve controllable watermark strength that balances between watermark strength and image quality, we propose a confidence-guided index replacement strategy. Specifically, we use the classification probability of an index predicted by the autoregressive model as the confidence measure and calculate relative confidence (will be detailed in Eq.~(\ref{eq:3})). The greater the relative confidence, the larger the gap between the red index and the paired green index at the current index position. Replacing these indices with significant gaps can lead to a noticeable decline in image quality. Based on the relative confidence distribution of two indices within each pair, we calculate a quantile as the replacement threshold to control watermark strength. For a given replacement threshold, we prioritize replacing index pairs with smaller relative confidence to balance watermark strength and image quality.

When the autoregressive model generates the $k$-th red index $\text{Idx}_k$, we record the classification probability $P(\text{Idx}_k)$ for $\text{Idx}_k$ and the classification probability $P(\text{Idx}_k')$ for its paired green index $\text{Idx}_k'$. After the autoregressive model generates all indices, we obtain the confidence set for red indices, 
$\text{conf} = \{P(\text{Idx}_1), P(\text{Idx}_2), \ldots, P(\text{Idx}_{N_{\text{red}}})\}$, 
and the confidence set for paired green indices, 
$\text{conf}' = \{P(\text{Idx}_1'), P(\text{Idx}_2'), \ldots, P(\text{Idx}_{N_{\text{red}}}')\}$, 
where $N_{\text{red}}$ represents the total number of red indices. Based on these two sets of confidence, we calculate the relative confidence for each index pair:
\begin{equation}
   \text{relative-conf}_k = \log(P(\text{Idx}_k)/P(\text{Idx}_k')),
   \label{eq:3}
\end{equation}
\begin{wrapfigure}[13]{r}{0.3\textwidth}
\centering
\vspace{-1em}
\includegraphics[width=0.3\textwidth]{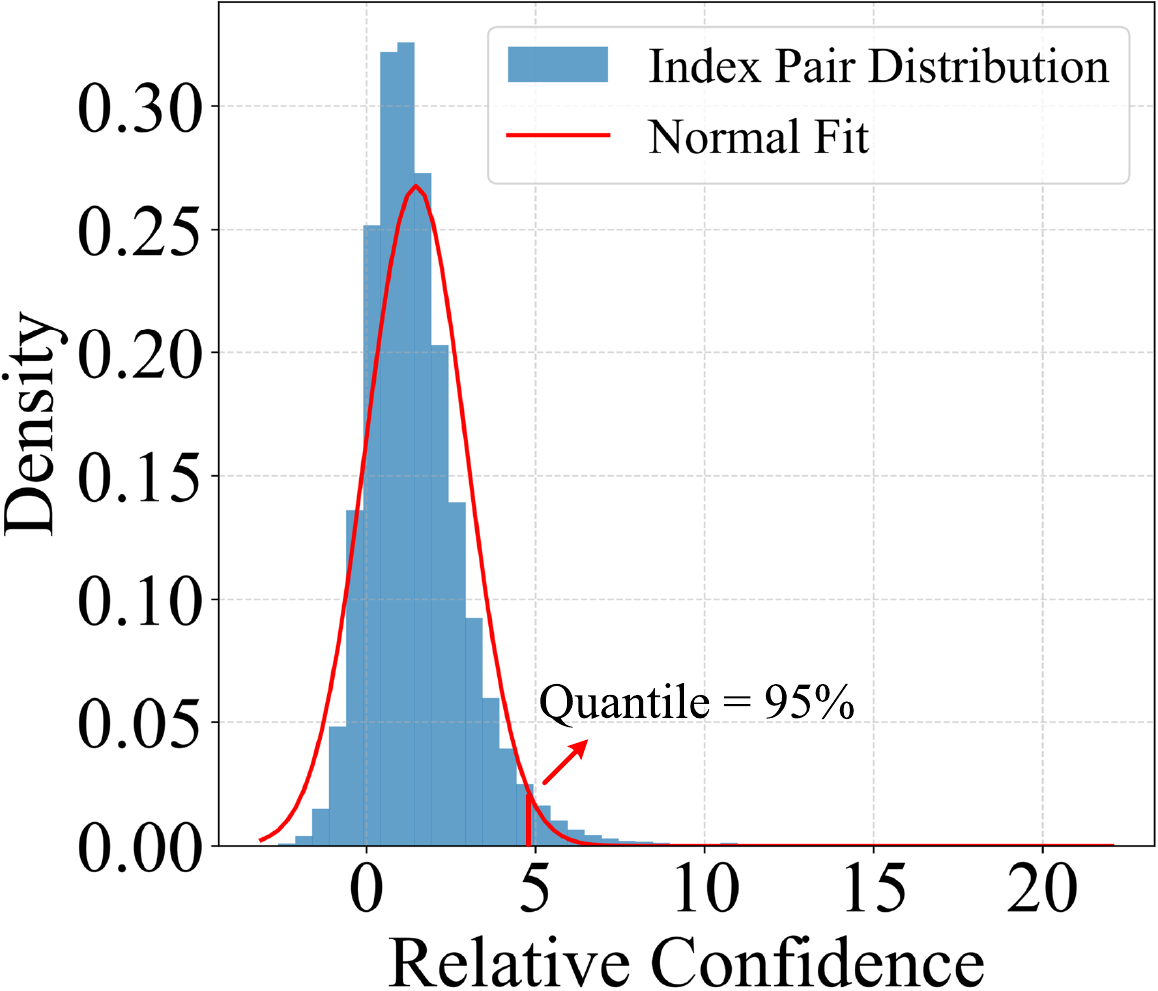}
\caption{\small
Index pair distribution of one hundred generated images.
}
\label{fig:hist}
\end{wrapfigure}
where $k$ represents the relative confidence of the $k$-th index pair. We achieve controllable watermark strength by setting a distribution quantile, with the relative confidence distribution illustrated in Figure~\ref{fig:hist}. Specifically, when replacing red indices with paired green indices, we only replace index pairs on the left side of the quantile. Therefore, when the quantile is set to 0\%, the model does not perform red index replacement, resulting in a non-watermarked image. When the quantile is set to 100\%, the model replaces all red indices with green indices from their index pairs. For other quantile values, the model prioritizes replacing red indices with green indices of lower relative confidence. Additionally, the confidence distribution exhibits a characteristic pattern of low density at both ends and high density in the middle, making index pairs with high relative confidence more likely to be filtered out. For example, by simply setting the quantile to 95\%, we can filter out all index pairs with a relative confidence greater than 5. This design not only achieves controllability of watermark strength but also optimizes the balance between watermark strength and image quality by ``filtering'' index pairs with high relative confidence.

\subsection{Watermark Verification}\label{sec:Watermark Verification}

\paragraph{Statistical Probability-Based Watermark Verification}

Since the red and green indices are randomly assigned, the proportion of green indices in an image without a watermark is approximately 50\%, while the proportion in an ideal watermarked image approaches 100\%. In fact, the process of autoregressive image generation can be regarded as $N_{\text{Idx}}$ independent Bernoulli trials with equal probability of taking the value 0 (red index) or 1 (green index), where $N_{\text{Idx}}$ is the total number of indices in an image. According to the Central Limit Theorem~\citep{clt}, when $N_{\text{Idx}}$ is sufficiently large, the sample mean follows a normal distribution. Thus, we can calculate the confidence interval CI for the mean of $N_{\text{Idx}}$ trials at a confidence level of $1-\beta$ as follows:
\begin{equation}
\text{CI} = \left(0.5 - \frac{z_{\beta/2}}{2\sqrt{N_{\text{Idx}}}}, \, 0.5 + \frac{z_{\beta/2}}{2\sqrt{N_{\text{Idx}}}}\right),
\end{equation}
where $z_{\beta/2}$ represents the two-tailed critical value of the standard normal distribution. After calculating the confidence interval, we can use the right endpoint of the confidence interval as a decision threshold. If the proportion of green indices in an image is below the threshold, the image is classified as a non-watermarked image; otherwise, it is classified as a watermarked image. 

\paragraph{Index Encoder}
\begin{figure*}[t]
    \centering
    \includegraphics[width=\linewidth]{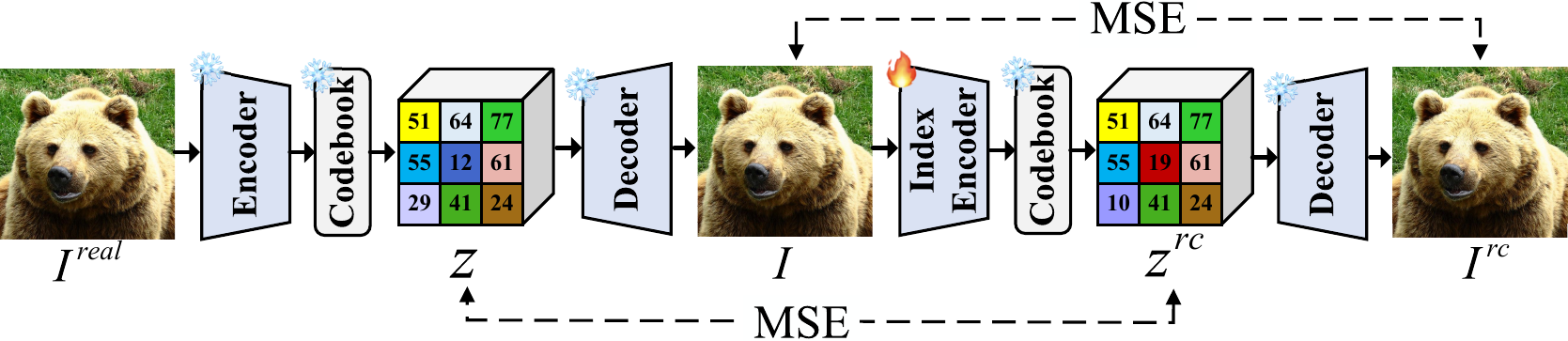}
    \caption{Training of Index Encoder. The Encoder, Codebook, and Decoder are frozen while the Index Encoder is updated to achieve accurate index reconstruction.}
    \label{fig:IE}
    \vspace{-15pt}
\end{figure*}
VQ-VAE is designed solely for pixel-level reconstruction. However, the objective of the watermark validator in this paper is index reconstruction. In practice, the autoregressively generated indices $\mathbf{Idx} = \{\text{Idx}_1, \text{Idx}_2, \ldots, \text{Idx}_{N_{\text{Idx}}}\}$ are processed through the decoder to produce the image $I$. 
However, the reconstructed indices $\mathbf{Idx}^{rc}$ obtained by feeding $I$ into the encoder and vector quantization module, may differ from the original $\mathbf{Idx}$. As illustrated in Figure~\ref{fig:point-b}, the original VQ-VAE encoder struggles to accurately identify watermarks in high-confidence scenarios. Therefore, we propose an \textit{Index Encoder} to assist users in high-confidence scenarios in achieving high-precision index reconstruction.

As shown in Figure~\ref{fig:IE}, we freeze the encoder, codebook, and decoder, and retrain a new encoder, termed \textit{Index Encoder}, with the goal of achieving accurate reconstruction of the indices $\mathbf{Idx}$. 
We first input the original image \(I^{real}\) into the encoder and decoder to obtain the vector \(z\) and image $I$. Then, we input $I$ into the Index Encoder and decoder to obtain the reconstructed vector \(z^{rc}\) and reconstructed image \(I^{rc}\). The optimization of the Index Encoder is performed by minimizing two loss terms: (1) the mean squared error (MSE) between the vector \(z\) and the vector \(z^{rc}\), and (2) the MSE between the image \(I\) and the reconstructed image \(I^{rc}\):

\begin{equation}
 \mathcal{L}_{encoder}= \| z^{rc} - z \|_2^2+\gamma  \|I^{rc} - I \|_2^2, 
\end{equation}
where $\gamma$ represents the weight hyperparameter.

\paragraph{Cropped Image Watermark Verification}

Although the red–green index watermark itself is highly robust, as demonstrated in Table~\ref{tab:quality}, the VQ-VAE encoding paradigm is inherently vulnerable to cropping attacks. During index reconstruction, VQ-VAE first divides an image into fixed-size, non-overlapping patches (\textit{e.g.}, $8\times8$ pixels) and independently encodes each patch to retrieve the index. 
Then, even a slight crop to the image can drastically alter the patch composition. For instance, consider cropping an image such that the new top-left corner lands at position (4, 3) within the neighborhood of the original patch. Such a tiny shift entails that every subsequent pixel now belongs to completely different spatial segments compared to the original image. When encoded, these reconfigured patches will lead to entirely different codebook indices, thereby significantly weakening watermark-verification robustness.

To address this weakness, we propose traversing every pixel in the local image block of the cropped image to achieve alignment of the local image blocks. Taking an $8\times8$ pixel block as an example, suppose the top-left corner of the cropped image is originally located at (4, 3) of a block. We enumerate the cropped image to traverse all pixel positions within the first local image block. That is, the top-left corner of the cropped image moves from (1, 1) to (8, 8), stopping after enumerating 64 candidate images. For each candidate, we calculate its green index rate. 
As long as the green index rate of one of the candidates reaches the decision threshold, the image is considered to contain a watermark. For example, as the image moves from (1, 1), the green index rate remains close to 50\%. However, when the top-left corner of the cropped image moves to (4, 3), the green index rate reaches 100\%, indicating the presence of a watermark. 
The example of cropped image verification can be found in the Appendix~\ref{sec:crop_example}.

\section{Experiments}\label{sec:Experiments}

\paragraph{Model and Datasets}
We conduct experiments using a state-of-the-art autoregressive image generation model, LlamaGen~\citep{llamagen}. For text-to-image generation tasks, we generate images at 256$\times$256 and 512$\times$512 resolutions. For class-conditioned image generation tasks, we generat images at 256$\times$256 and 384$\times$384 resolutions. We conduct pre-training for index reconstruction at various resolutions on LlamaGen's text-to-image VQ-VAE and class-conditioned VQ-VAE using the MS-COCO-2017 training dataset~\citep{mscoco} and the ImageNet-1k validation dataset~\citep{deng2009imagenet}, respectively.

\paragraph{Evaluation Metrics and Baselines}
To evaluate the effectiveness of IndexMark, we employ the watermark verification method based on statistical probability, calculating accuracy (ACC) to measure the watermark verification performance. In addition, we utilize Peak Signal-to-Noise Ratio (PSNR), Structural Similarity Index (SSIM), and Multiscale SSIM (MSSIM)~\citep{wang2004image} to quantify the pixel-level differences between watermarked and original images. We further assess the fidelity of the watermarked image distribution with the FID~\citep{heusel2017gans} and evaluate the alignment between generated images and their text prompts with the CLIP score~\citep{radford2021learning}. We compare IndexMark with five baseline approaches: three post-processing methods including DwtDct~\citep{dwtdct}, DwtDctSvd~\citep{navas2008dwt}, and RivaGAN~\citep{RivaGAN}, and two diffusion-based methods including Tree-Ring~\citep{tree} and ROBIN~\citep{robin}.
\footnote{More details about how the evaluation metrics are computed can be found in the Appendix~\ref{Implementation Details}.}

\paragraph{Implementation Details}\label{sec:Implementation Details}
For the construction of index pairs, we set top-K pruning with $K=10$. For the text-conditioned generation task, we set top-K sampling with $K=1000$, CFG-scale to 7.5, and downsample-size to 16. For the class-conditioned generation task, we set top-K sampling with $K=2000$, CFG-scale to 4.0, downsample-size to 16, and default to a full-green index watermark. For the Index Encoder, we used the Adam optimizer~\citep{adam} with a learning rate of 1e‐5, and set $\gamma$ to 0.5. All experiments are conducted on an NVIDIA A100 GPU.

\begin{table*}[t]
\centering
\caption{Comparison of IndexMark with post- and in-processing watermarking methods in terms of quality and robustness against various attacks.}
\resizebox{\textwidth}{!}{
\begin{tabular}{cc|ccccc|ccccccc|c}
\toprule
\multirow{2}{*}{Model} & \multirow{2}{*}{Method}&\multicolumn{5}{c|}{Image quality}& &\multicolumn{5}{c}{Accuracy $\uparrow$} &    \\ && PSNR $\uparrow$ & SSIM $\uparrow$ & MSSIM $\uparrow$  & CLIP $\uparrow$   & FID$\downarrow$ & Clean & Blur & Noise & JPEG & Bright & Erase & Crop & Avg \\
\midrule
\multicolumn{15}{c}{\textbf{MSCOCO Dataset}} \\
\midrule
     \multirow{3}{*}{
     \shortstack[c]{Post processing \\ (256 $\times$ 256)}} 
   & DwtDct & 37.71 & 0.970   & \textbf{0.992}  &  0.325  &25.85 &  0.603 & 0.501 &     0.607  &    0.500  &0.571 &  0.567&0.500&0.549 \\
   & DwtDctSvd &37.57  &  0.979    & \textbf{0.992}  &  0.325  &27.60& 0.996  &0.982  & 0.994      & 0.963     & 0.556& 0.994 &0.500&0.855\\
   & RivaGAN   & \textbf{40.44} & \textbf{0.980} & \textbf{0.992} &  0.324& 25.78  & 0.930  &0.919  &   0.929    &0.727     & 0.862& 0.847 & 0.500&0.816 \\
   
\cdashline{1-15}
   
     \multirow{2}{*}{
     \shortstack[c]{LlamaGen (AR) \\ (256 $\times$ 256)}}  & \cellcolor{gray!10}\textcolor{gray}{W/o watermark} & \cellcolor{gray!10}\textcolor{gray}{$\infty$} & \cellcolor{gray!10}\textcolor{gray}{1.000} & \cellcolor{gray!10}\textcolor{gray}{1.000} & \cellcolor{gray!10}\textcolor{gray}{0.328} & \cellcolor{gray!10}\textcolor{gray}{26.55} &  
     \cellcolor{gray!10}\textcolor{gray}{$-$} & 
     \cellcolor{gray!10}\textcolor{gray}{$-$} &
     \cellcolor{gray!10}\textcolor{gray}{$-$} &
     \cellcolor{gray!10}\textcolor{gray}{$-$} &
     \cellcolor{gray!10}\textcolor{gray}{$-$} &
     \cellcolor{gray!10}\textcolor{gray}{$-$} &
     \cellcolor{gray!10}\textcolor{gray}{$-$} &
    \cellcolor{gray!10}\textcolor{gray}{$-$}\\
   & IndexMark    & 23.54 & 0.838 & 0.930 & \textbf{0.326} & $\textbf{24.73}$ & \textbf{1.000}  &\textbf{0.991}  & \textbf{0.995}    & \textbf{0.978}   &\textbf{0.988} &\textbf{0.997} &  \textbf{0.998}&\textbf{0.992} \\

   \cmidrule{1-15}
   
     \multirow{3}{*}{
     \shortstack[c]{Post processing \\ (512 $\times$ 512)}}
   & DwtDct & 37.61 & 0.963   & \textbf{0.990}  &   0.279 & 54.30& 0.741  &0.512  &  0.739     &0.500      &0.680 & 0.734 &0.500&0.629\\
    & DwtDctSvd & 37.38 & 0.972   & 0.989  &  0.280  &55.60&0.999  & 0.990 &   \textbf{0.998}    & 0.988     &0.673 & 0.998&0.500&0.878\\
    & RivaGAN    & \textbf{40.41} & \textbf{0.978} & 0.989 & 0.279 &   56.49& 0.973 & 0.967 &   0.970    &   0.900   &0.930 &  0.945&0.958&0.949   \\

\cdashline{1-15}
   
     \multirow{3}{*}{
     \shortstack[c]{Stable Diffusion \\ (512 $\times$ 512)}}& \cellcolor{gray!10}\textcolor{gray}{W/o watermark} & \cellcolor{gray!10}\textcolor{gray}{$\infty$} & \cellcolor{gray!10}\textcolor{gray}{1.000} & \cellcolor{gray!10}\textcolor{gray}{1.000} & \cellcolor{gray!10}\textcolor{gray}{0.403} & \cellcolor{gray!10}\textcolor{gray}{25.53} & 
     \cellcolor{gray!10}\textcolor{gray}{$-$} & 
     \cellcolor{gray!10}\textcolor{gray}{$-$} &
     \cellcolor{gray!10}\textcolor{gray}{$-$} &
     \cellcolor{gray!10}\textcolor{gray}{$-$} &
     \cellcolor{gray!10}\textcolor{gray}{$-$} &
     \cellcolor{gray!10}\textcolor{gray}{$-$} &
     \cellcolor{gray!10}\textcolor{gray}{$-$} &
    \cellcolor{gray!10}\textcolor{gray}{$-$}\\
   & Tree-Ring& $15.37$ & $0.568$   & $0.626$  & $0.364$   & $\textbf{25.93}$ &\textbf{1.000} & \textbf{1.000} &  0.994     &   \textbf{0.999}   &\textbf{1.000} &\textbf{1.000}  &0.833&0.975\\
   & ROBIN    & ${24.03}$ & ${0.768}$ & ${0.881}$ & $\textbf{0.396}$ & $26.86$ & \textbf{1.000}  &  \textbf{1.000}&    \textbf{0.998}   &   0.971   &\textbf{1.000} & \textbf{1.000} &0.918 &0.983 \\

\cdashline{1-15}
   
    \multirow{2}{*}{
     \shortstack[c]{LlamaGen (AR) \\ (512 $\times$ 512)}} & \cellcolor{gray!10}\textcolor{gray}{W/o watermark} & \cellcolor{gray!10}\textcolor{gray}{$\infty$} & \cellcolor{gray!10}\textcolor{gray}{1.000} & \cellcolor{gray!10}\textcolor{gray}{1.000} & \cellcolor{gray!10}\textcolor{gray}{0.282} & \cellcolor{gray!10}\textcolor{gray}{54.57} &  
     \cellcolor{gray!10}\textcolor{gray}{$-$} & 
     \cellcolor{gray!10}\textcolor{gray}{$-$} &
     \cellcolor{gray!10}\textcolor{gray}{$-$} &
     \cellcolor{gray!10}\textcolor{gray}{$-$} &
     \cellcolor{gray!10}\textcolor{gray}{$-$} &
     \cellcolor{gray!10}\textcolor{gray}{$-$} &
     \cellcolor{gray!10}\textcolor{gray}{$-$} &
    \cellcolor{gray!10}\textcolor{gray}{$-$}\\
   & IndexMark    & 24.15 & 0.838 & 0.930 & 0.281 & 54.35   &\textbf{1.000}  &0.988  & 0.994      &   0.984   &0.989 & 0.992 &\textbf{0.993}&\textbf{0.991} \\

\midrule 
\multicolumn{15}{c}{\textbf{ImageNet Dataset}} \\
\midrule 
     \multirow{3}{*}{
     \shortstack[c]{Post processing \\ (256 $\times$ 256)}} 
   & DwtDct & 38.73 & 0.974   & \textbf{0.993}  & $\textbf{0.288}$   & 15.17 &   0.583   &  0.501   & 0.588 &     0.500   &   0.584      & 0.568    &  0.500    &0.546\\
   & DwtDctSvd & 38.44 & 0.979   & 0.991  & $\textbf{0.288}$   & 15.32&  0.994    &  0.991   & 0.989 &     0.960   &    0.552     &    0.994 &0.500   &0.854 \\
   & RivaGAN   & \textbf{40.44} & \textbf{0.980} & 0.991 & $\textbf{0.288}$ & 15.29 &   0.951    &  0.930   &  0.950&    0.746    &  0.919       & 0.914    &  0.500  & 0.844  \\

\cdashline{1-15}

    \multirow{3}{*}{
     \shortstack[c]{ImageNet Diffusion \\ (256 $\times$ 256)}} & \cellcolor{gray!10}\textcolor{gray}{W/o watermark}   & \cellcolor{gray!10}\textcolor{gray}{$\infty$} & \cellcolor{gray!10}\textcolor{gray}{1.000} & \cellcolor{gray!10}\textcolor{gray}{1.000} & \cellcolor{gray!10}\textcolor{gray}{$0.271$} & \cellcolor{gray!10}\textcolor{gray}{16.25}    &  
     \cellcolor{gray!10}\textcolor{gray}{$-$} & 
     \cellcolor{gray!10}\textcolor{gray}{$-$} &
     \cellcolor{gray!10}\textcolor{gray}{$-$} &
     \cellcolor{gray!10}\textcolor{gray}{$-$} &
     \cellcolor{gray!10}\textcolor{gray}{$-$} &
     \cellcolor{gray!10}\textcolor{gray}{$-$} &
     \cellcolor{gray!10}\textcolor{gray}{$-$} &
    \cellcolor{gray!10}\textcolor{gray}{$-$}\\
 & Tree-Ring& $15.68$   & $0.663$          & $0.607$          & $0.267$         & ${17.68}$      &   \textbf{1.000}  & 0.994 &0.999       &  0.998    & 0.798&  0.995& 0.924&0.958         \\
  & ROBIN    & ${24.98}$ & ${0.875}$ & ${0.872}$ & $0.275$ & $18.26$&     \textbf{1.000} & 0.999 &0.999       & 0.999     &0.928 &0.999  &  0.994&0.988   \\

\cdashline{1-15}

     \multirow{2}{*}{
     \shortstack[c]{LlamaGen (AR) \\ (256 $\times$ 256)}} & \cellcolor{gray!10}\textcolor{gray}{W/o watermark} & \cellcolor{gray!10}\textcolor{gray}{$\infty$} & \cellcolor{gray!10}\textcolor{gray}{1.000} & \cellcolor{gray!10}\textcolor{gray}{1.000} & \cellcolor{gray!10}\textcolor{gray}{$0.289$} & \cellcolor{gray!10}\textcolor{gray}{15.08}  & 
     \cellcolor{gray!10}\textcolor{gray}{$-$} & 
     \cellcolor{gray!10}\textcolor{gray}{$-$} &
     \cellcolor{gray!10}\textcolor{gray}{$-$} &
     \cellcolor{gray!10}\textcolor{gray}{$-$} &
     \cellcolor{gray!10}\textcolor{gray}{$-$} &
     \cellcolor{gray!10}\textcolor{gray}{$-$} &
     \cellcolor{gray!10}\textcolor{gray}{$-$} &
    \cellcolor{gray!10}\textcolor{gray}{$-$}\\


   & IndexMark    & ${23.86}$ & ${0.738}$ & ${0.903}$ & $\textbf{0.288}$ & \textbf{13.89} & \textbf{1.000}  &  \textbf{1.000}&  \textbf{1.000}     &   \textbf{1.000}   &\textbf{0.998} & \textbf{1.000} &\textbf{0.998} &\textbf{0.999} \\

\cmidrule{1-15}
     \multirow{3}{*}{
     \shortstack[c]{Post processing \\ (384 $\times$ 384)}}  
   & DwtDct & 39.36 & 0.972   & \textbf{0.991}  & $\textbf{0.286}$   & 12.50& 0.720 &0.521  &   0.725    &   0.500   & 0.780&0.696  &0.500&0.634 \\
   & DwtDctSvd & 39.08 & \textbf{0.979}   & 0.989  & $0.285$   & 12.62& 0.999 & 0.990 &  0.999     &   0.542   &0.664 & 0.999 & 0.500&0.813 \\
   & RivaGAN    & \textbf{40.45} & 0.977 & 0.989 & $\textbf{0.286}$ & 12.79 & 0.966  & 0.947 &    0.964   & 0.846     &0.949 &  0.999& 0.953 &0.946  \\

\cdashline{1-15}
     \multirow{2}{*}{
     \shortstack[c]{LlamaGen (AR) \\ (384 $\times$ 384)}}& \cellcolor{gray!10}\textcolor{gray}{W/o watermark} & \cellcolor{gray!10}\textcolor{gray}{$\infty$} & \cellcolor{gray!10}\textcolor{gray}{1.000} & \cellcolor{gray!10}\textcolor{gray}{1.000} & \cellcolor{gray!10}\textcolor{gray}{$0.287$} & \cellcolor{gray!10}\textcolor{gray}{12.65}  & 
     \cellcolor{gray!10}\textcolor{gray}{$-$} & 
     \cellcolor{gray!10}\textcolor{gray}{$-$} &
     \cellcolor{gray!10}\textcolor{gray}{$-$} &
     \cellcolor{gray!10}\textcolor{gray}{$-$} &
     \cellcolor{gray!10}\textcolor{gray}{$-$} &
     \cellcolor{gray!10}\textcolor{gray}{$-$} &
     \cellcolor{gray!10}\textcolor{gray}{$-$} &
    \cellcolor{gray!10}\textcolor{gray}{$-$}\\
   & IndexMark    & ${25.45}$ & ${0.783}$ & ${0.913}$ & $\textbf{0.286}$ & \textbf{11.81}& \textbf{1.000}  &\textbf{1.000}  &   \textbf{1.000}    & \textbf{1.000}     &\textbf{0.998} & \textbf{1.000} & \textbf{0.993} &\textbf{0.998}  \\

\bottomrule
\end{tabular}
}
\label{tab:quality}
\vspace{-10pt}
\end{table*}

\subsection{Image Quality and Watermark Robustness}
\begin{figure*}[t]
    \centering
    \includegraphics[width=0.9\linewidth]{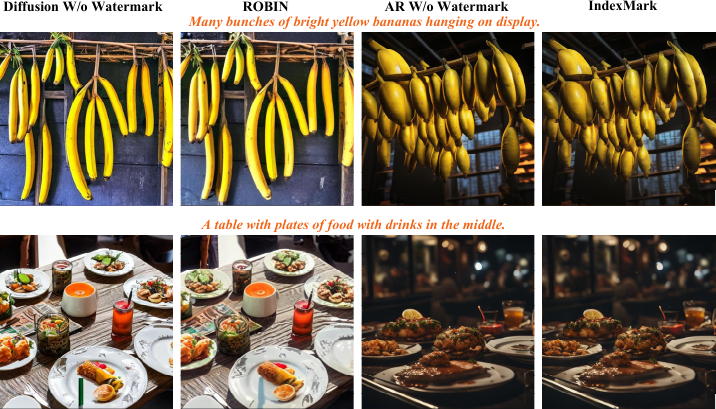}
    \caption{ROBIN vs.~IndexMark. ROBIN embeds watermarks during the intermediate diffusion state, which may lead to changes in the image content. In contrast, IndexMark uses the \textit{match-then-replace} strategy to embed watermarks, effectively preserving the image's quality and content.}
    \label{fig:robinwar}
    \vspace{-10pt}
\end{figure*}

\paragraph{Image Quality}
Traditional post-hoc watermarking methods often cause slight visual distortions and suffer from poor robustness. In contrast, generative methods can seamlessly embed watermarks into the generated content without altering the semantics. Diffusion-based methods typically operate in the latent spacebut tend to cause significant semantic changes due to the difficulty in precisely controlling the perturbation magnitude. In comparison, our approach is built upon VQ-VAE and autoregressive image generation models and therefore performs better in faithfully preserving image details and structures. As shown in Table \ref{tab:quality}, our approach achieves significant improvements in the PSNR, SSIM, and MSSIM scores, while causing much less image quality degradation compared to watermark-free generations, as evidenced by the CLIP and FID scores. Moreover, unlike diffusion-based methods, we observe that the FID of watermarked images in our method is even lower than that of non-watermarked ones, further demonstrating our superiority in preserving visual fidelity. Figure~\ref{fig:robinwar} shows that the ROBIN method may remove certain objects from the original image (such as chopsticks and pastries on the plate), which affects parts of the generated image content. In contrast, our method effectively preserves the overall image content and semantic structure, demonstrating its superiority. More qualitative results can be found in the Appendix~\ref{Qualitative Results}.

\paragraph{Robustness}
To evaluate the robustness of our watermarking method, we select six common data augmentations as attack methods. These include Gaussian blur with a kernel size of 11, Gaussian noise with a standard deviation of $\sigma=0.01$, JPEG compression with a quality factor of 70, color jitter with brightness set to 0.5, random erasing of 10\% of the region, and random cropping of 75\%. We select the right endpoint of the confidence interval at a 99.9\% confidence level as the threshold for watermark determination. As shown in Table \ref{tab:quality}, we report the ACC under each attack setting. Notably, our method demonstrates strong robustness against most perturbations, significantly outperforming the baselines at image resolutions of 256, 384, and 512. While Stable Diffusion-based methods perform better than traditional approaches, they still fall noticeably short of our method.

\subsection{Ablation Study and Further Analyses}
\begin{figure}[t]
  \centering
  \subfigure[Evaluation on confidence-guided index replacement.]{\includegraphics[width=0.45\linewidth]{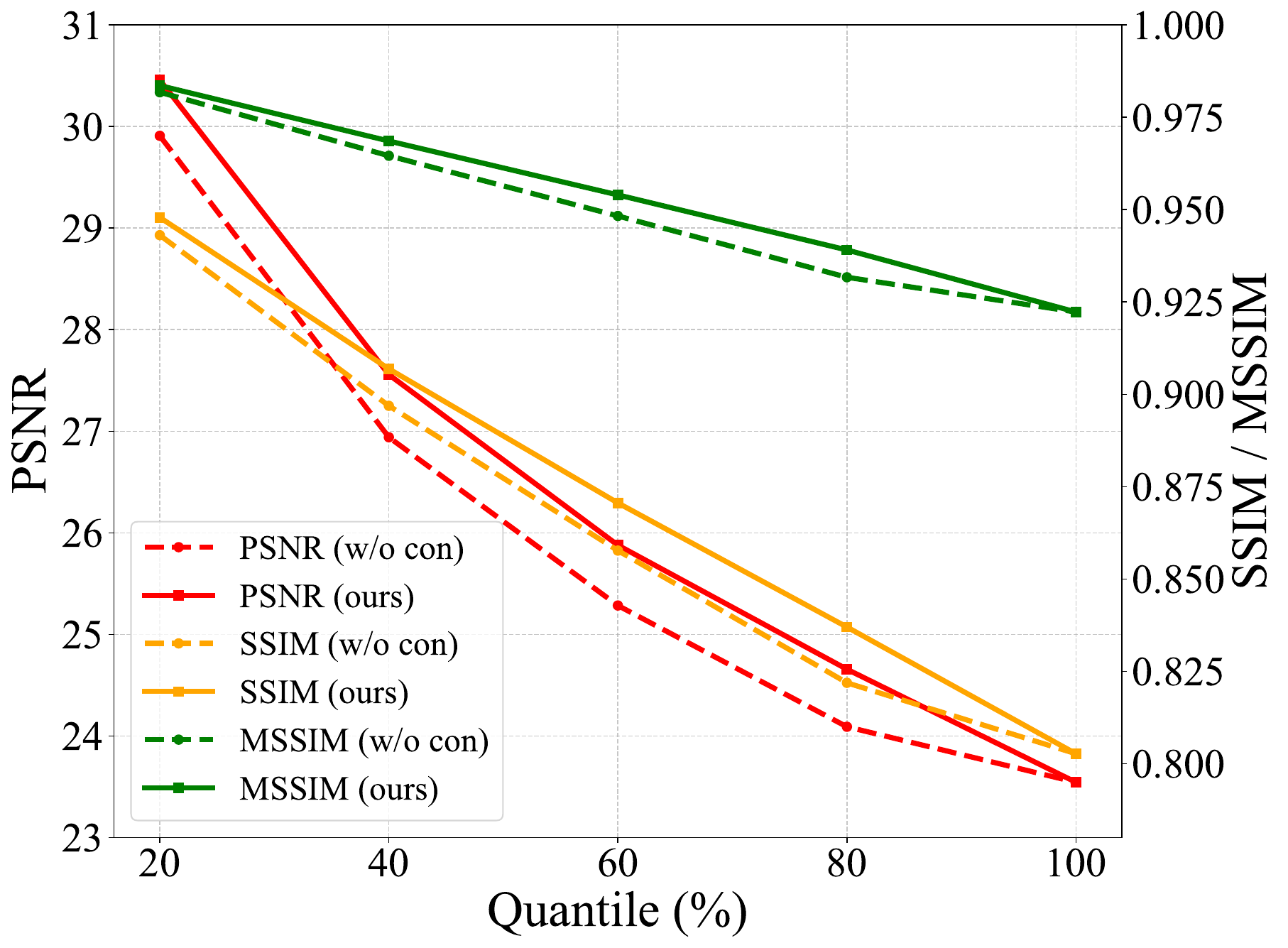}
    \label{fig:point-a}}
  \subfigure[Evaluation on Index Encoder.]{\includegraphics[width=0.45\linewidth]{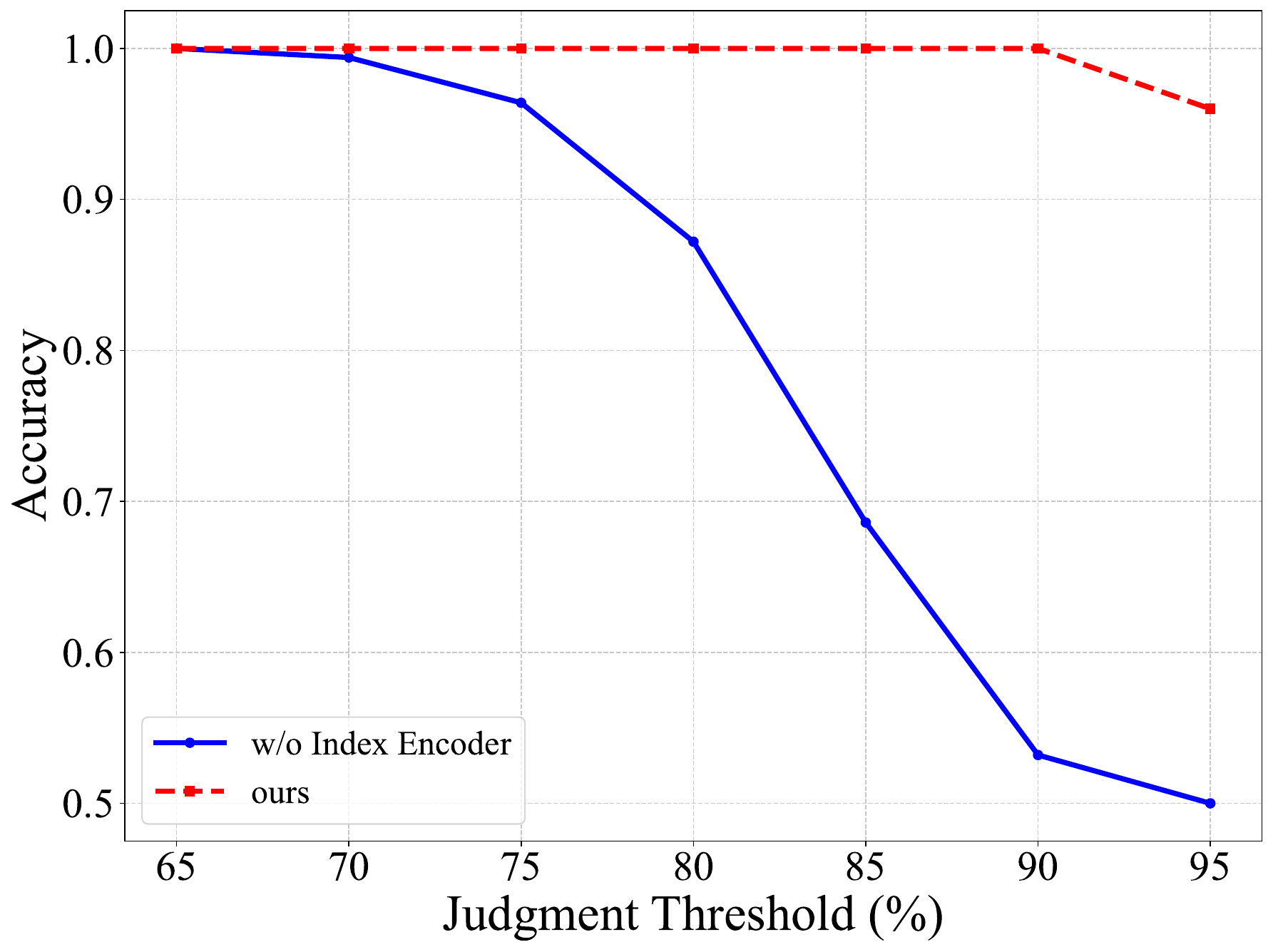}
    \label{fig:point-b}}
  
  \caption{Ablation results on confidence-guided index replacement and Index Encoder.}
  \label{fig:combined}
  
  \vspace{-0.6cm}
\end{figure}

\paragraph{Confidence-Guided Index Replacement}
We substitute the confidence-guided method with random index selection based on watermark strength. The results shown in Figure~\ref{fig:point-a} justify the effectiveness of our design as the PSNR scores of random index selection are significantly lower than IndexMark.

\paragraph{Index Encoder}
We compare watermark verification rates with and without Index Encoder at different confidence levels on 256$\times$256 resolution images. As shown in Figure~\ref{fig:point-b}, differences are minimal at lower confidence levels, but at higher levels, Index Encoder significantly improves verification rates. More robustness experiments can be found in the Appendix~\ref{robie}.

\begin{figure}[t]
  \centering
  \subfigure[Training Loss.]{\includegraphics[width=0.3\linewidth]{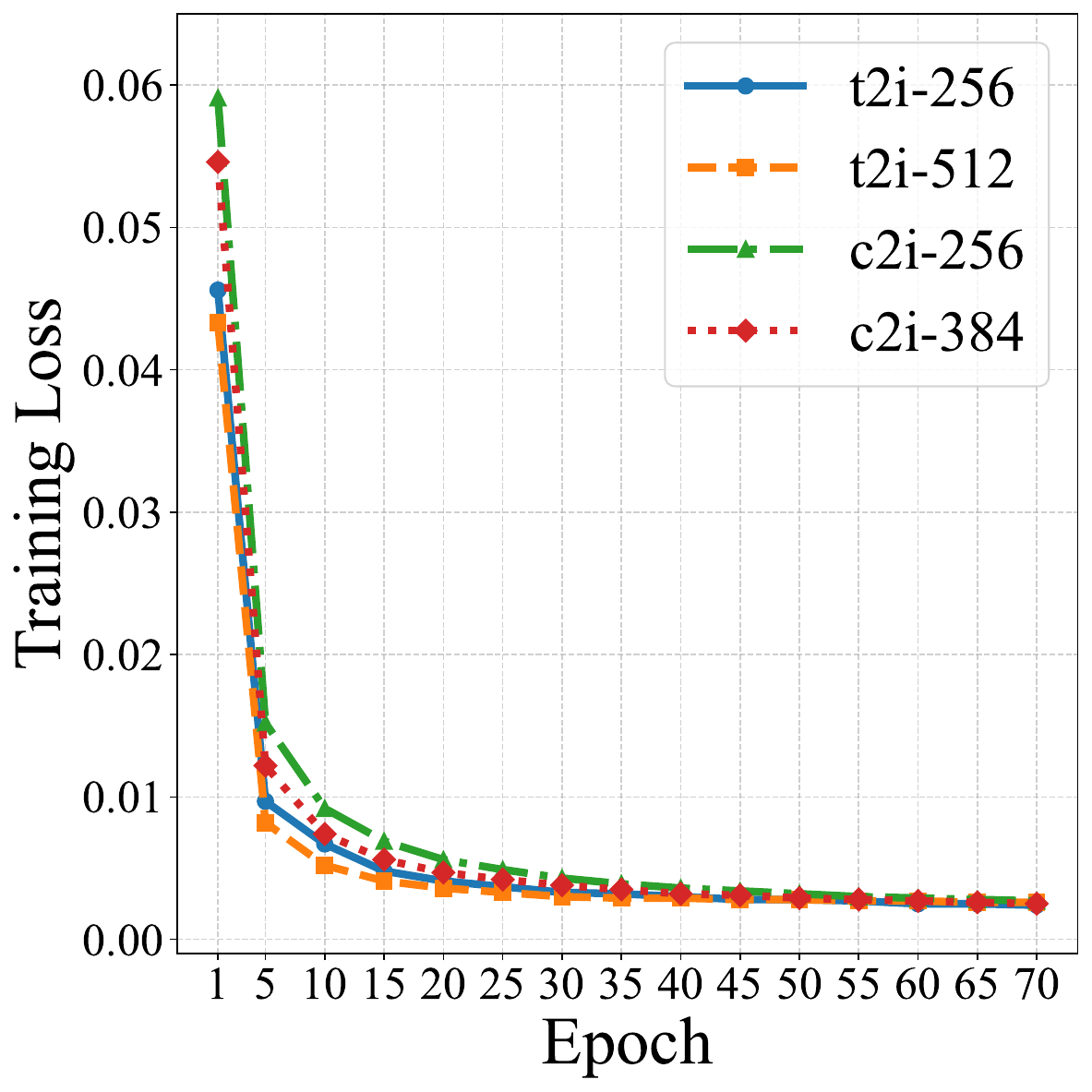}\label{fig:Indexencoder-a}}
  \subfigure[Index Reconstruction Rate.]{\includegraphics[width=0.3\linewidth]{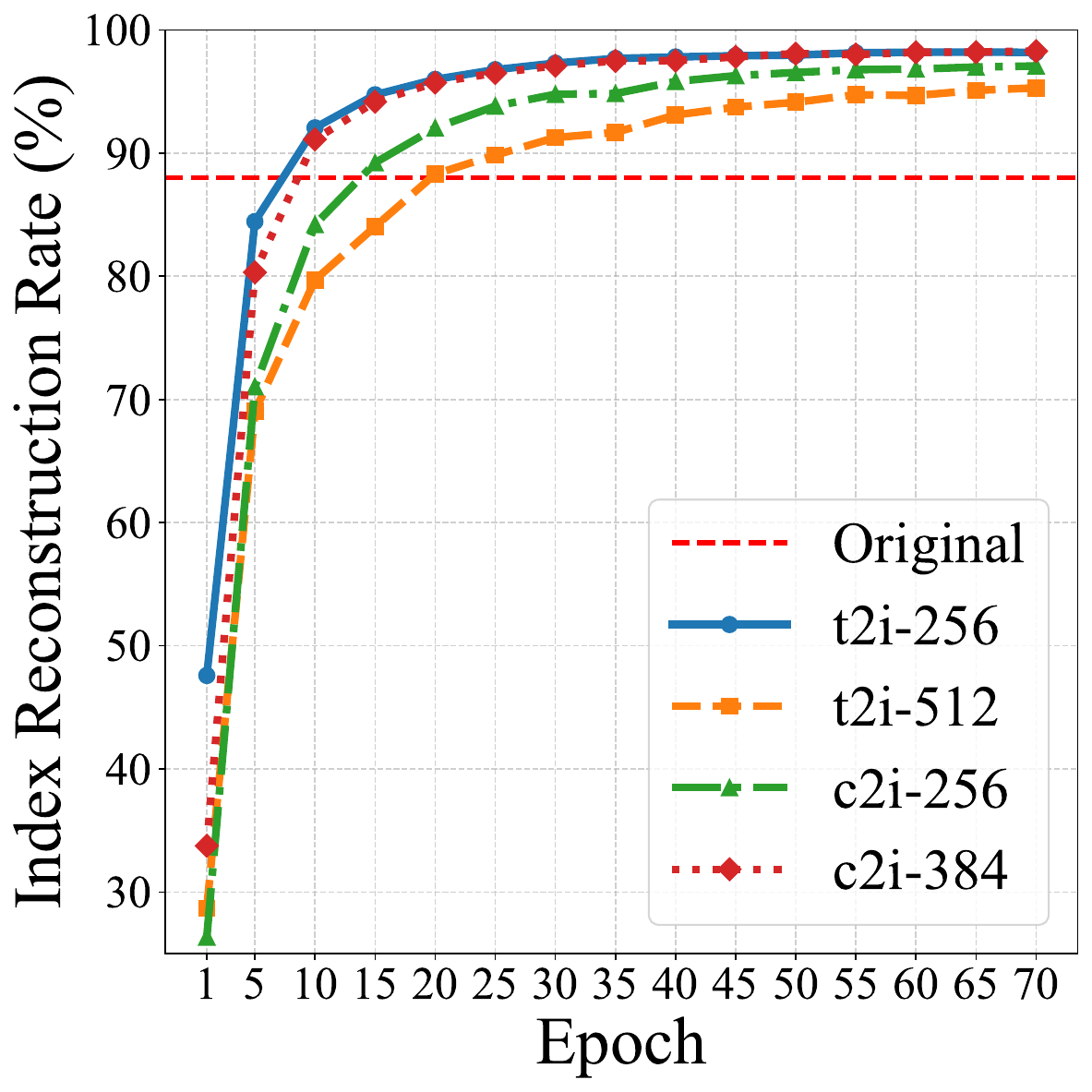}\label{fig:Indexencoder-b}}
  \subfigure[Green index Validation Rate.]{\includegraphics[width=0.3\linewidth]{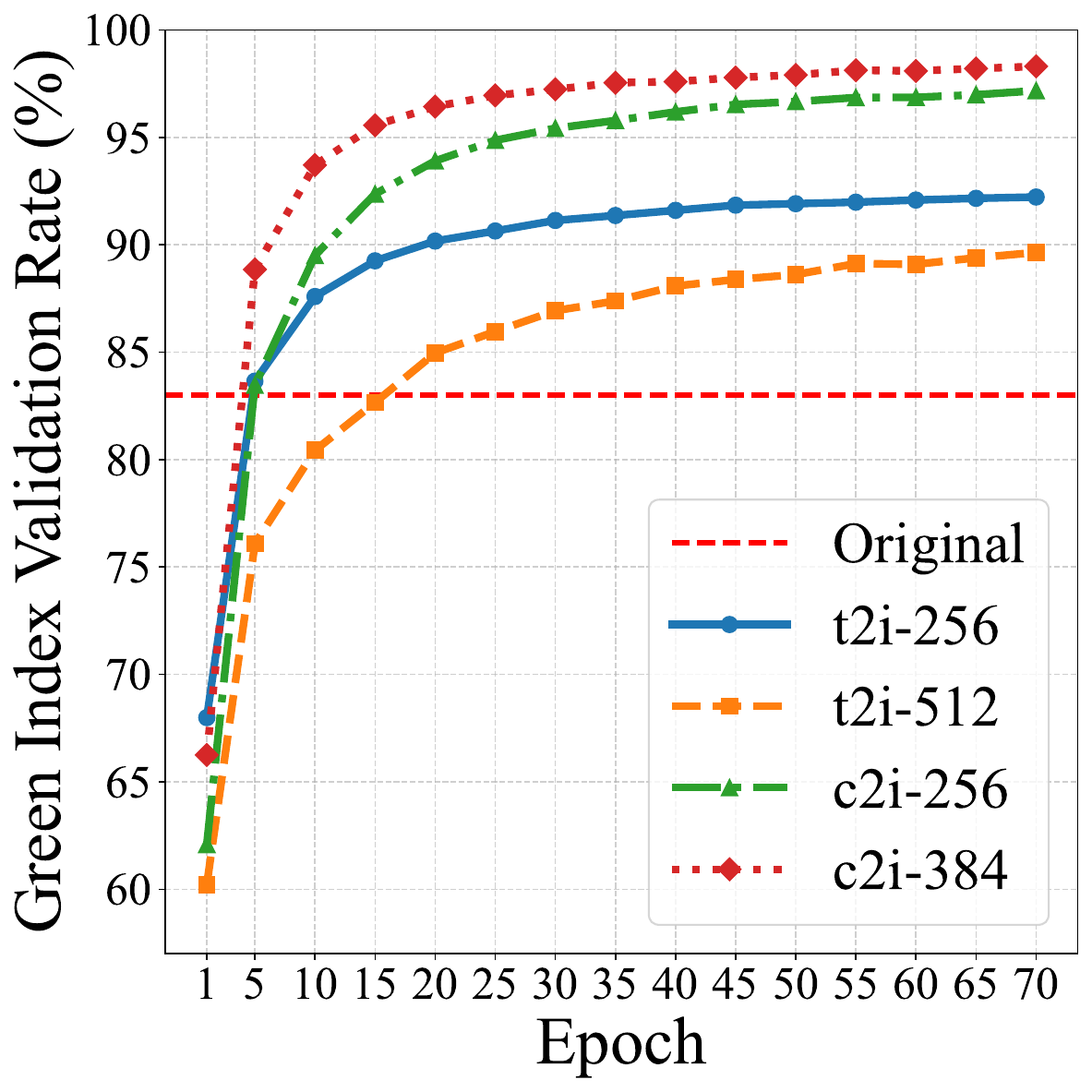}\label{fig:Indexencoder-c}}
  \caption{Images showing the variation of training loss, index reconstruction rate, and green index verification rate of the Index Encoder with respect to epochs.}
  \label{fig:Indexencoder_train}
  \vspace{-0.5cm}
\end{figure}

\paragraph{Index Reconstruction}
To investigate whether the Index Encoder improves index reconstruction capability, we conduct index-to-index reconstruction experiments at a resolution of 256 across multiple epochs using the Index Encoder, as well as validation experiments on pure green index images. Additionally, Figure~\ref{fig:Indexencoder-a} illustrates the training loss across multiple resolutions. As shown in Figure~\ref{fig:Indexencoder-b}, after only 20 epochs of training, the index reconstruction capability of the Index Encoder surpasses that of the original encoder. Furthermore, as depicted in Figure~\ref{fig:Indexencoder-c}, the Index Encoder's validation capability for images with all indices being green significantly exceeds that of the original encoder, allowing users to verify watermarks with a higher confidence level.

\begin{figure*}[t]
    \centering
    \includegraphics[width=0.9\linewidth]{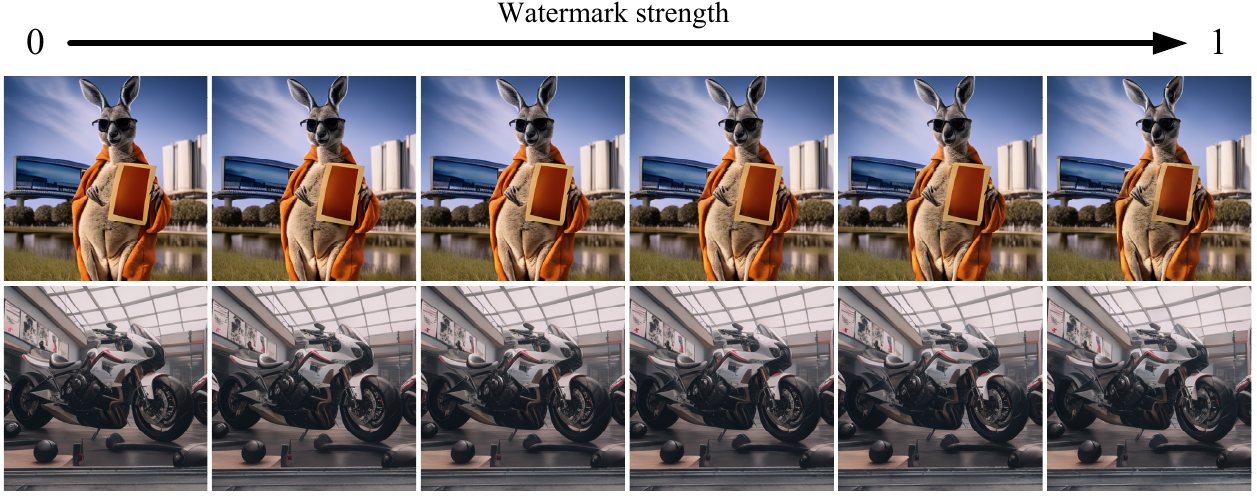}
    \caption{Generated images under different watermark strengths.}
    \label{fig:qualitative results}
    \vspace{-10pt}
\end{figure*}

\paragraph{Watermark Strength} We explore the impact of watermark strength on images. The qualitative results, as shown in Figure~\ref{fig:qualitative results}, indicate that an increase in IndexMark watermark strength does not cause noticeable changes in image quality.

\section{Conclusion}
This paper proposes IndexMark, the first \textit{training-free} watermarking method for autoregressive image generation models. IndexMark carefully selects watermark tokens from the codebook based on token similarity and promotes the use of watermark tokens through token replacement, thereby embedding the watermark in the image. We believe that our method offers a novel perspective for watermark design in autoregressive image generation models.

\bibliographystyle{plain}
\bibliography{reference}
\clearpage
\appendix

\section*{Appendix} \label{appendix}
\section{Limitations and Social Impact}
\subsection{Limitations} \label{sec:limitations}
The verification of IndexMark watermark relies on the index reconstruction capability of the VQ-VAE model. A more robust encoder can enhance the robustness of our method, such as index reconstruction based on image semantics~\citep{1dvqvae}. Additionally, our current match-then-replace method uses simple pairwise matching. By exploring diverse matching methods, we can further leverage the redundancy of the codebook, thereby improving the quality of the watermarked images.
\subsection{Social Impact}\label{sec:Social Impact}
With the rapid advancement of autoregressive image generation models, developers have the responsibility and obligation to ensure the safety of these models. We provide developers with an efficient and effective method to help them counteract the misuse of models, marking a step towards responsible AI in autoregressive image generation models.
\section{Model Details}
\subsection{VQ-VAE and Autoregressive Image Generation}
\label{vqvaear}
\paragraph{VQ-VAE} 
The Vector Quantized Variational Autoencoder (VQ-VAE) provides a framework for encoding images into a discrete latent representation. Given an input image \( x \in \mathbb{R}^{H \times W \times 3}\), the encoder produces a continuous latent feature map: 
\begin{equation}
z = \text{encoder}(x)\in \mathbb{R}^{h \times w \times d}.
\end{equation}
For every spatial location $(i,j)$ we find the nearest entry in the VQ-VAE's codebook \( \mathcal{C} = \{e_1, e_2, \dots, e_K\} \subset\mathbb{R}^d\):
\begin{equation}
    k_{ij}
              \;=\;
              \operatorname*{arg\,min}_{k\in\{1,\dots,K\}}
              \lVert z_{ij} - e_k\rVert_2,
              \qquad
              z_{ij}^q \;=\; e_{k_{ij}},
\end{equation}
where \(k_{ij}\) is a discrete index, and \(z_{ij}^q\) is the corresponding quantised vector. By flattening the quantized vector $z^{q}$, a sequence of discrete tokens \(T= \{T_1, T_2, \dots, T_{h \times w}\} \) is obtained, where each token \( T_i \) represents an index in the codebook \( \mathcal{C} \).  
During the reconstruction phase, the quantized latent vector \( z^q \) is retrieved using the token indices and the codebook. This vector is then passed through a decoder to reconstruct the original image: $ \hat{x} = \text{decoder}(z^q) $. During the training phase, the model is constrained by the image reconstruction loss, codebook loss, and commitment loss, defined as:
\begin{equation}
    \mathcal{L} = \|x - \hat{x}\|_2^2 + \beta \|q - \text{sg}[z]\|_2^2 + \gamma \|z - \text{sg}[q]\|_2^2,
    \label{eq:vqvae_loss}
\end{equation}
where $\text{sg}$ denotes the stop gradient operation.
\paragraph{Autoregressive Image Generation} The autoregressive model defines the generation process as the prediction of the next token:
\begin{equation}
p(\mathbf{x})
= \prod_{i=1}^n p\bigl(x_i \mid x_1, x_2, \dots, x_{i-1}\bigr)
= \prod_{i=1}^n p(x_i \mid x_{<i}).
\end{equation}
In autoregressive image generation, $x_i$ denotes the image token in the discrete latent space, and the image generation process can be formulated as:
\begin{equation}
p(\mathbf{q})
= \prod_{i=1}^{h\times w} p(q_i \mid q_{<i},c),
\end{equation}
where $q_i$ denotes the discretized image token, $c$ denotes the embedding of the
class label or the text, and $h \times w$ represents the total number of image tokens. During the training phase, the model is trained by maximizing the likelihood of the observed token sequences:
\begin{equation}
L_{\mathrm{train}}
= -\log p(\mathbf{q})
= -\sum_{i=1}^{h\times w} \log p\bigl(q_i \mid q_{<i},c\bigr).
\end{equation}
During inference, the model generates the sequence of token indices autoregressively by sampling each next index. Once the full sequence of image token indices is produced, the codebook is used to reconstruct the latent vector $z_q$ from those indices, and $z_q$ is then fed into the VQ-VAE decoder to synthesize the final image.

\subsection{Blossom}
\label{blossomapp}
The core principle of the Blossom algorithm is to iteratively approach the optimal matching by dynamically handling odd-length cycle structures within the graph. Its key steps are as follows:
\begin{itemize}
    \item \textbf{Blossom Shrinking}: When the algorithm verifies an odd cycle, it contracts the cycle into a super vertex, preserving the connections between the cycle and external vertices, thereby simplifying the complex structure into a recursively manageable subgraph.
    \item \textbf{Augmenting Path Search}: The current matching is expanded by traversing a path that alternates between matched and unmatched edges. During each expansion, the matching status of the edges on the path is flipped to increase the total weight.
    \item \textbf{Dual Variable Adjustment}: Utilizing the duality theory of linear programming, the potentials of vertices and odd sets are adjusted to ensure that each operation converges toward maximizing the total weight.
\end{itemize}
The pseudocode of the Blossom Algorithm is shown in Algorithm~\ref{blossom}.

\begin{algorithm}[H]
\DontPrintSemicolon
\KwIn{Graph $G = (V, E)$, edge weights $w: E \to \mathbb{R}$}
\KwOut{Maximum-weight perfect matching $M \subseteq E$}

\tcp{Initialize}
$M \gets \varnothing$ \;
$y(v) \gets \frac{1}{2} \max_{e \in \delta(v)} w(e)$ for all $v \in V$ \;
$\mathcal{B} \gets \varnothing$ \;

\While{$M$ is not perfect}{
    Search for augmenting paths via BFS/DFS \tcp*{Build alternating trees}

    \If{any odd-length cycle $B$ found}{
        \tcp{Blossom Shrinking}
        Contract $B$ into super-node $b$ \;
        Update $\mathcal{B} \gets \mathcal{B} \cup \{b\}$ \;
        Adjust dual variables $y$ and $z_B$ for $b$ \tcp*{Maintain LP feasibility}
    }

    \If{augmenting path $P$ found}{
        \tcp{Augment matching}
        $M \gets M \oplus P$ \tcp*{Symmetric difference} \;
        Expand blossoms in $\mathcal{B}$ along $P$ \tcp*{Restore original graph} \;
        Reset search structures \;
    }

    \tcp{Dual Variable Adjustment}
    Compute $\delta = \min\{\text{slack}(e) \mid e \in E\}$ \;
    Update $y(v) \gets y(v) \pm \delta$ and $z_B \gets z_B + 2\delta$ \tcp*{Converge to optimality}
}
\Return{$M$}
\caption{Blossom Algorithm}
\label{blossom}
\end{algorithm}

\subsection{Watermark Verification on Cropped Image}\label{sec:crop_example}
\begin{figure*}[ht]
    \centering
    \includegraphics[width=\linewidth]{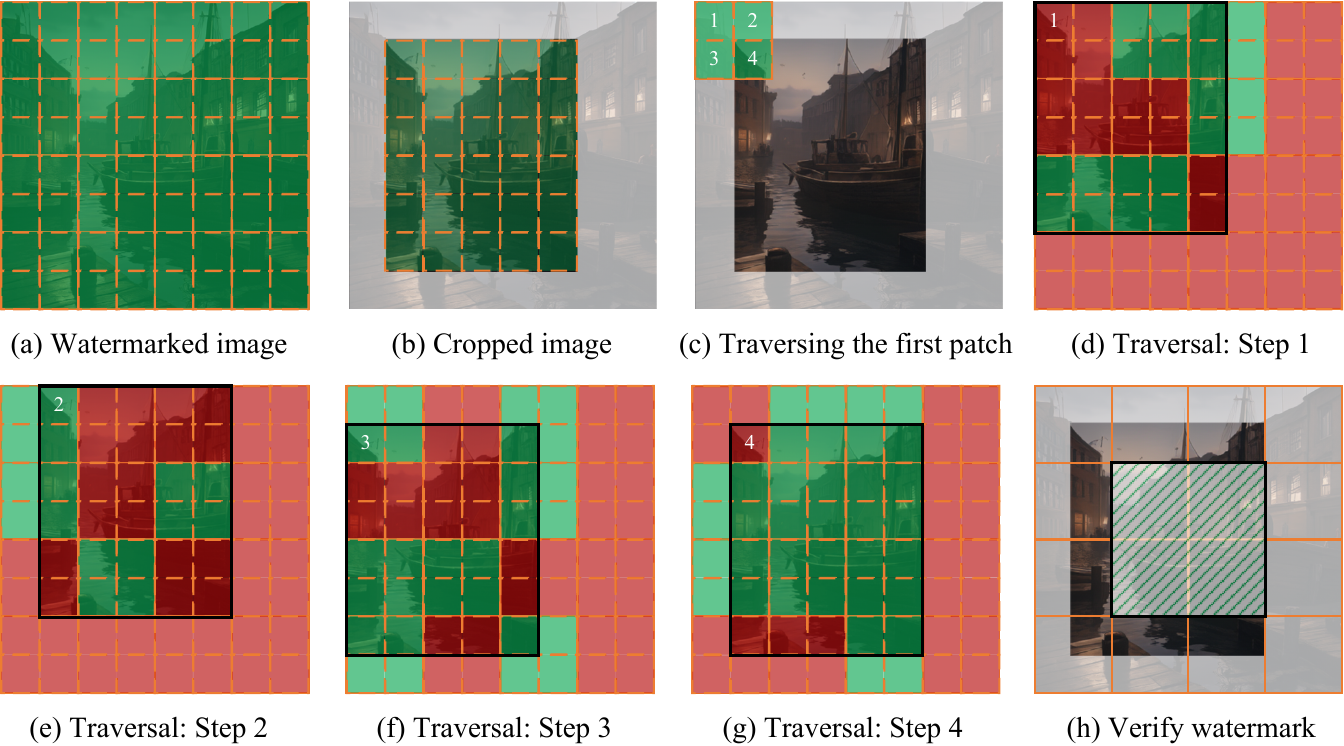}
    \caption{Visualization of the traversal process for watermark verification on the cropped image.}
    \label{fig:crop_var}
\end{figure*}

In the Figure~\ref{fig:crop_var}, we show watermark verification process on a cropped image. As an example, with an $8\times8$ input image and using a patch side length of 2, by traversing the first image patch, the watermark in the cropped image can be successfully verified with at most $2\times2$ checks.

\section{Experimental Details}
\label{Implementation Details}
\subsection{Details About Evaluation Metrics}
\paragraph{FID} For text-to-image tasks, we generate 5,000 images to evaluate the Fréchet Inception Distance (FID) score~\citep{heusel2017gans} on the MS-COCO-2017 training dataset. For class-conditioned image generation tasks, we generate 10,000 images to evaluate the FID score on the ImageNet-1k validation dataset.
\paragraph{CLIP Score} We use OpenCLIP-ViT model~\citep{clipscore} to compute the CLIP score~\citep{radford2021learning} between generated images and their corresponding text prompts. For class-conditioned generation, we use ``a photo of category'' as the input. 

\subsection{Details of the Threshold for Watermark Determination}
For $256 \times 256$ and $384 \times 384$ resolutions, we select a green index rate of 0.615 near the 99.9\% confidence level as the determination threshold. For $512 \times 512$ resolution, we choose a green index rate of 0.60 near the 99.99\% confidence level as the determination threshold. Regarding cropping attacks, since the image is reduced to approximately 50\% of its original size, we use a higher confidence level to detect the watermark. Specifically, for $512 \times 512$ resolution, we use a green index rate of 0.65 as the determination threshold, while for $256 \times 256$ and $384 \times 384$ resolutions, we adopt 0.7 as the determination threshold.

\section{More Experimental Results}
\subsection{Green Index Generation}
\label{greenexperiment}
\begin{figure*}[ht]
    \centering
    \includegraphics[width=0.8\linewidth]{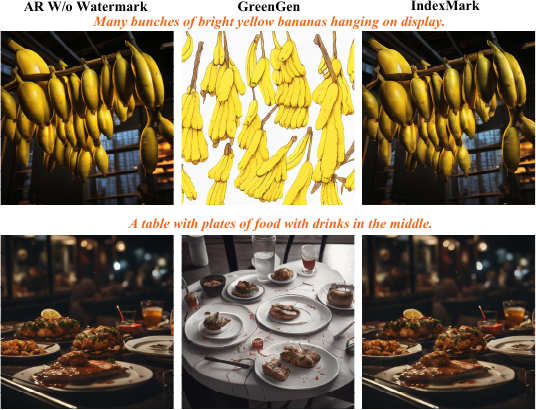}
    \caption{GreenGen vs.~IndexMark. GreenGen generates autoregressive images by removing red indices from the codebook and using only green indices, resulting in significant differences between the watermarked images and non-watermarked ones. In contrast, IndexMark achieves smaller differences through a match-then-replace method.}
    \label{fig:GreenGen}
\end{figure*}

We explored the possibility of generating images using only the green indices from the codebook, referring to this variant as GreenGen. As shown in Figure~\ref{fig:GreenGen}, the watermarked images generated by GreenGen exhibit significant differences compared to the watermark-free images. The quantitative results are shown in Table~\ref{tab:greenindex}. GreenGen differs significantly from the watermarked image at the pixel level. Although GreenGen achieves a CLIP score similar to that of IndexMark, its performance in terms of FID is not as good as IndexMark. This result indicates that there is a substantial amount of redundancy in the codebook, and our method effectively leverages this redundancy to achieve better watermark embedding while maintaining image quality and content integrity.

\begin{table}[ht]
\centering
\caption{Comparison results of image quality between IndexMark and GreenGen.}
\scalebox{0.99}{
\begin{tabular}{cccccccc}
\toprule
Model & Method    & PSNR $\uparrow$ & SSIM $\uparrow$ & MSSIM $\uparrow$  & CLIP $\uparrow$   & FID $\downarrow$   \\
\midrule
\multicolumn{7}{c}{\textbf{MSCOCO Dataset}} \\
\midrule
    \multirow{3}{*}{
     \shortstack[c]{LlamaGen (AR) \\ (256 $\times$ 256)}}  & \cellcolor{gray!10}\textcolor{gray}{W/o watermark} & \cellcolor{gray!10}\textcolor{gray}{$\infty$} & \cellcolor{gray!10}\textcolor{gray}{1.000} & \cellcolor{gray!10}\textcolor{gray}{1.000} & \cellcolor{gray!10}\textcolor{gray}{0.328} & \cellcolor{gray!10}\textcolor{gray}{26.55}   \\
   & GreenGen & $9.76$ &  $0.267$  & $0.111$  &  \textbf{0.326}  &26.35  \\
   & IndexMark    & \textbf{23.54} & \textbf{0.838} & \textbf{0.930} & \textbf{0.326} & $\textbf{24.73}$   \\
\midrule
    \multirow{3}{*}{
     \shortstack[c]{LlamaGen (AR) \\ (512 $\times$ 512)}} & \cellcolor{gray!10}\textcolor{gray}{W/o watermark} & \cellcolor{gray!10}\textcolor{gray}{$\infty$} & \cellcolor{gray!10}\textcolor{gray}{1.000} & \cellcolor{gray!10}\textcolor{gray}{1.000} & \cellcolor{gray!10}\textcolor{gray}{0.282} & \cellcolor{gray!10}\textcolor{gray}{54.57}   \\
   & GreenGen & $10.11$ &  $0.280$  & $0.129$  & \textbf{0.281}  &54.51  \\
   & IndexMark    & \textbf{24.15} & \textbf{0.838} & \textbf{0.930} & \textbf{0.281} & \textbf{54.35}   \\
\midrule
\multicolumn{7}{c}{\textbf{ImageNet Dataset}} \\
\midrule
 \multirow{3}{*}{
      \shortstack[c]{LlamaGen (AR) \\ (256 $\times$ 256)}} & \cellcolor{gray!10}\textcolor{gray}{W/o watermark} & \cellcolor{gray!10}\textcolor{gray}{$\infty$} & \cellcolor{gray!10}\textcolor{gray}{1.000} & \cellcolor{gray!10}\textcolor{gray}{1.000} & \cellcolor{gray!10}\textcolor{gray}{$0.289$} & \cellcolor{gray!10}\textcolor{gray}{15.08}   \\

 & GreenGen & $9.46$ &  $0.186$ & $0.106$  & $\textbf{0.288}$   &15.30  \\

 & IndexMark    & $\textbf{23.86}$ & $\textbf{0.738}$ & $\textbf{0.903}$ & $\textbf{0.288}$ & \textbf{13.89}   \\
\midrule

    \multirow{3}{*}{
     \shortstack[c]{LlamaGen (AR) \\ (384 $\times$ 384)}}& \cellcolor{gray!10}\textcolor{gray}{W/o watermark} & \cellcolor{gray!10}\textcolor{gray}{$\infty$} & \cellcolor{gray!10}\textcolor{gray}{1.000} & \cellcolor{gray!10}\textcolor{gray}{1.000} & \cellcolor{gray!10}\textcolor{gray}{$0.287$} & \cellcolor{gray!10}\textcolor{gray}{12.65}   \\
   & GreenGen & $9.454$ &  $0.230$  & $0.131$  & $\textbf{0.286}$   & 12.46 \\
   & IndexMark    & $\textbf{25.45}$ & $\textbf{0.783}$ & $\textbf{0.913}$ & $\textbf{0.286}$ & \textbf{11.81}   \\
\bottomrule
\end{tabular}
}
\label{tab:greenindex}
\end{table}
\subsection{Robustness Experiment without the Index Encoder}\label{robie}
Under lower watermark‐verification confidence thresholds, we removed the Index Encoder (w/o IE) and conducted robustness experiments. As shown in Table~\ref{tab:accwoindexencoder}, even at low confidence settings, the model without the Index Encoder maintains strong robustness, thereby reducing training costs for users with less stringent security requirements.

\begin{table*}[ht]
    \caption{Comparison of ACC across different watermarking methods under various attacks.
Clean indicates watermark verification results on unaltered images, while Avg represents the average accuracy across all attack scenarios.}
    \centering
    \scalebox{0.9}{
    \begin{tabular}{ccccccccc|c}
         \toprule
        {Model} & {Method}      & Clean & Blur & Noise & JPEG & Bright & Erase & Crop & Avg \\
\midrule
\multicolumn{10}{c}{\textbf{MSCOCO Dataset}} \\
\midrule
\multirow{2}{*}{\begin{tabular}[c]{@{}c@{}}LlamaGen (AR) \\ (256 $\times$ 256)\end{tabular}}
&IndexMark w/o IE   &\textbf{1.000}  &0.972  & 0.990      & 0.970     & 0.974& \textbf{0.997} &0.917 &0.974  \\
&IndexMark    &\textbf{1.000}  &\textbf{0.991}  & \textbf{0.995}    & \textbf{0.978}   &\textbf{0.988} &\textbf{0.997} &  \textbf{0.998}&\textbf{0.992} \\
\midrule
\multirow{2}{*}{\begin{tabular}[c]{@{}c@{}}LlamaGen (AR) \\ (512 $\times$ 512)\end{tabular}}
&IndexMark w/o IE   &\textbf{1.000}  &0.969 &  0.992    & \textbf{0.980}    &0.981 &0.992  & \textbf{0.939}&0.978  \\
&IndexMark   & \textbf{1.000}  &  \textbf{1.000}&    \textbf{0.998}   &  0.971   &\textbf{1.000} & \textbf{1.000} &0.918 &\textbf{0.983}  \\ 
\midrule
\multicolumn{10}{c}{\textbf{ImageNet Dataset}} \\
\midrule
\multirow{2}{*}{\begin{tabular}[c]{@{}c@{}}LlamaGen (AR) \\ (256 $\times$ 256)\end{tabular}}
&IndexMark w/o IE    &\textbf{1.000}   & \textbf{1.000} &   \textbf{1.000}    &    \textbf{1.000}  &0.995 & \textbf{1.000} & 0.996&0.998 \\
&IndexMark   &\textbf{1.000}  &  \textbf{1.000}&  \textbf{1.000}     &   \textbf{1.000}   &\textbf{0.998} & \textbf{1.000} & \textbf{0.998} &\textbf{0.999} \\
\midrule
\multirow{2}{*}{\begin{tabular}[c]{@{}c@{}}LlamaGen (AR) \\ (384 $\times$ 384)\end{tabular}}
&IndexMark w/o IE    &\textbf{1.000}  & 0.999 &  0.999     &   \textbf{1.000}  &0.994 & 0.999 & 0.905&0.985  \\ 
&IndexMark   &\textbf{1.000}  &\textbf{1.000}  &   \textbf{1.000}    & \textbf{1.000}     &\textbf{0.998} & \textbf{1.000} & \textbf{0.993} &\textbf{0.998} \\ 
\bottomrule
\end{tabular}}
\label{tab:accwoindexencoder}
\end{table*}

\newpage
\subsection{More Qualitative Results}
\label{Qualitative Results}
\begin{figure*}
    \centering
    \includegraphics[width=0.9\linewidth]{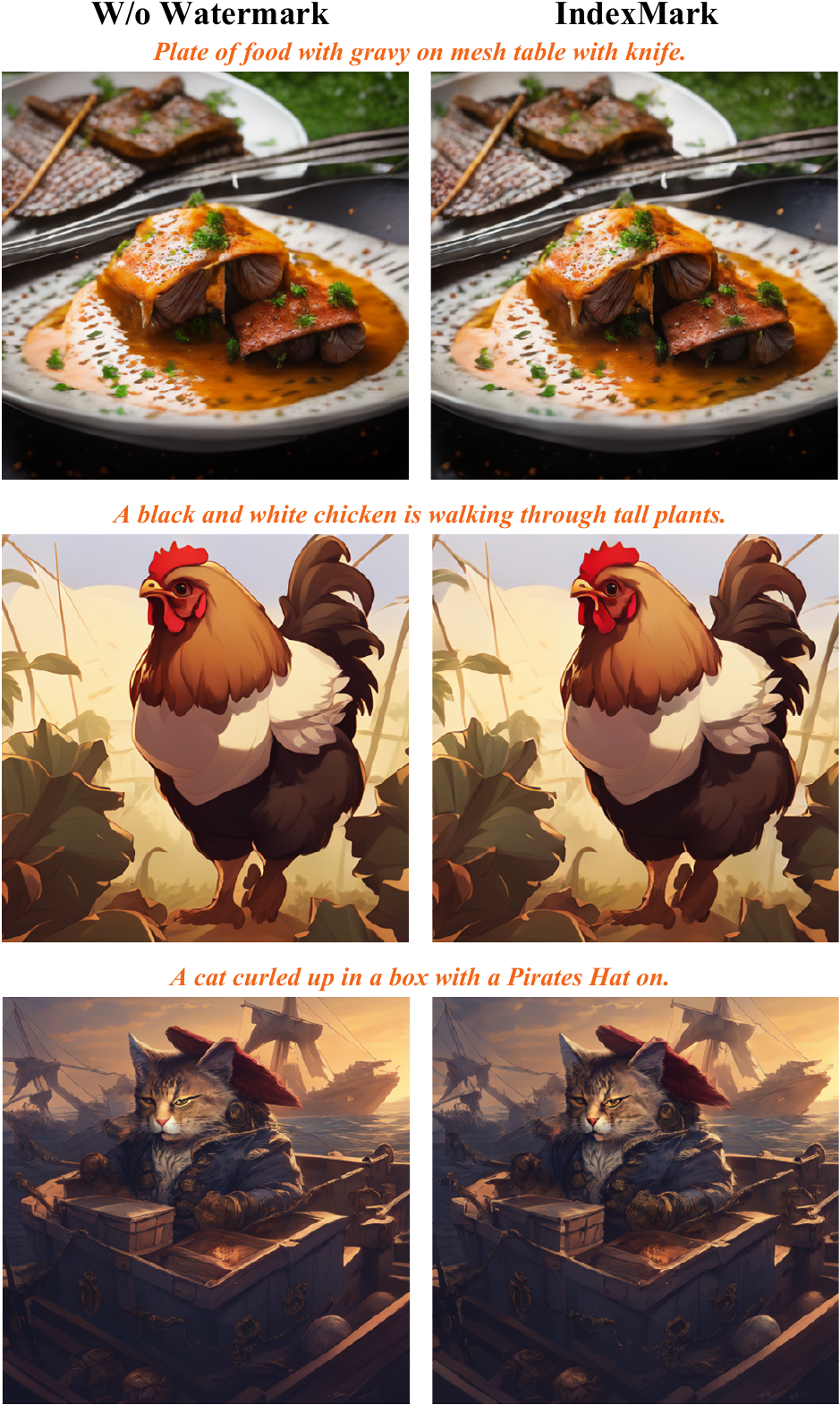}
    \caption{More qualitative comparison results between non-watermarked images and IndexMark watermarked images.}
    \label{fig:append1}
\end{figure*}
\begin{figure*}
    \centering
    \includegraphics[width=0.9\linewidth]{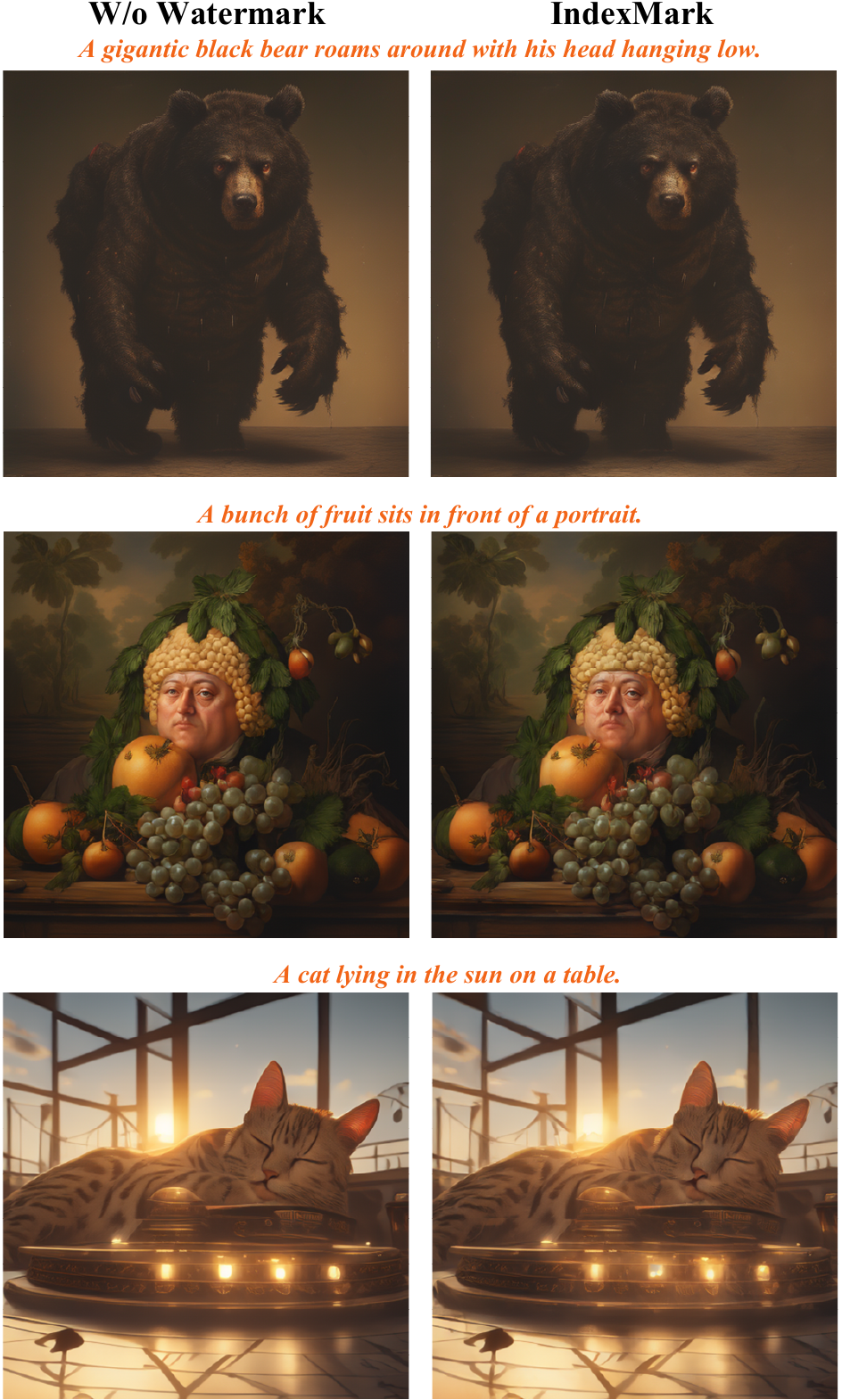}
    \caption{More qualitative comparison results between non-watermarked images and IndexMark watermarked images.}
    \label{fig:append2}
\end{figure*}
\end{document}